\documentclass{article}

     \usepackage[preprint]{neurips_2024}
\usepackage{amsmath}
\usepackage{graphicx}

\usepackage[utf8]{inputenc} 
\usepackage[T1]{fontenc}    
\usepackage{hyperref}       
\usepackage{url}            
\usepackage{booktabs}       
\usepackage{amsfonts}       
\usepackage{nicefrac}       
\usepackage{microtype}      
\usepackage{xcolor}         

\title{Assessing the Robustness of Climate Foundation Models under No-Analog Distribution Shifts}

\author{%
  Maria Conchita Agana Navarro \\
  Centre for Artificial Intelligence\\
  Department of Computer Science\\
  University College London, UK\\
   \And
   Geng Li \\
  Division of Emerging Interdisciplinary Areas\\
Hong Kong University of Science and Technology \\
Hong Kong\\
   \And
   Theo Wolf \\
   Department of Computer Science\\
   University of Oxford, UK \\
   \And
   Mar\'ia P\'erez-Ortiz \\
   Centre for Artificial Intelligence\\
  Department of Computer Science\\
  University College London\\
  \texttt{maria.perez@ucl.ac.uk} \\
}

\begin{document}

\maketitle

\begin{abstract}
The accelerating pace of climate change introduces profound non-stationarities that challenge the ability of Machine Learning (ML)-based climate emulators to generalize beyond their training distributions. While these emulators offer computationally efficient alternatives to traditional Earth System Models (ESMs), their reliability remains a {potential} bottleneck under ``no-analog'' future climate states---{which we define here as regimes where external forcing drives the system into conditions outside the empirical range of the historical training data}. {A fundamental challenge in evaluating this reliability is data contamination; because many models are trained on simulations that already encompass future scenarios, true out-of-distribution (OOD) performance is often masked. To address this, we} systematically benchmark the OOD robustness of three state-of-the-art architectures---U-Net, ConvLSTM, and the ClimaX foundation model---{specifically restricted to a} historical-only training regime (1850--2014). We evaluate these models using two complementary strategies: (i) temporal extrapolation to the recent climate (2015--2023) and (ii) cross-scenario forcing shifts across divergent emission pathways (SSP1-2.6 and SSP5-8.5).
Our analysis {within this controlled experimental setup} reveals {an accuracy vs. stability trade-off}: while the ClimaX foundation model achieves the lowest absolute error, it exhibits {higher relative performance changes} under distribution shifts, with precipitation errors increasing by up to 8.44\% under extreme forcing scenarios. Conversely, simpler CNN-based architectures demonstrate {greater relative stability}. Furthermore, we identify a temperature-precipitation {disparity}, where models maintain stable temperature projections while exhibiting larger degradation in precipitation under forcing shifts. These findings suggest that {when restricted to historical training dynamics, even high-capacity foundation models} are sensitive to external forcing trajectories. Our results underscore the {necessity of} scenario-aware training and rigorous OOD evaluation protocols that account for data contamination to ensure the robustness of climate emulators under a changing climate.
\end{abstract}

\section{Introduction}

The accelerating pace of climate change and the increasing severity of extreme weather events present urgent challenges for global adaptation and mitigation planning. The 2022 IPCC Sixth Assessment Report (AR6) highlights that continued warming will amplify climate extremes and heighten the risk of crossing critical system thresholds, such as Amazon rainforest dieback or irreversible ice-sheet loss \citep{RN4, IPCC2023}. As governments and industries design infrastructure, assess financial risks, and plan for diverse future emission trajectories, the need for reliable climate projections has never been more pressing.

Global Climate Models (GCMs) remain the cornerstone of climate projection, providing physically based simulations of the {coupled atmosphere–ocean–land system \citep{IPCC2023}}. Yet, their computational cost limits the number of scenarios, ensembles, and high-resolution experiments that can be produced, constraining the exploration of uncertainty across socioeconomic pathways and climate trajectories.

Machine Learning (ML) offers an appealing complement to traditional modeling: ML-based climate emulators can approximate GCM outputs orders of magnitude faster, enabling rapid experimentation and scenario exploration \citep{kaltenborn2023climatesetlargescaleclimatemodel}. 
Recent advances---from convolutional and recurrent architectures to {Foundation models such as ClimaX \citep{nguyen2023climaxfoundationmodelweather} have demonstrated strong skill, yet recent critiques suggest they may overfit to internal variability, leading to performance trade-offs compared to simpler linear baselines \citep{lutjens2025impact}. Furthermore, the ability of these models to maintain fidelity when transferred across divergent forcing scenarios remains an open scientific issue \citep{addison2024machine, kendon2025potential}.}
 {These approaches generally address two distinct settings: forcing-to-climate mapping (e.g., ClimateBench \citep{https://doi.org/10.1029/2021MS002954}) and weather-learning systems that attempt to derive climate statistics from short-term dynamics, e.g. NeuralGCM \citep{kochkov2024neural} or ACE \citep{watt2025ace2}). The latter represents a significantly more complex task, as it requires the model to learn weather physics and accurately aggregate them into stable climate states.} However, these gains mask a core unresolved challenge: \textit{the ability of ML models to generalize under non-stationary climate conditions}.

Unlike many ML applications, the climate system is evolving under strong external forcing, meaning that future states will increasingly diverge from the historical distributions on which most ML models are trained. {This raises a fundamental methodological challenge regarding data contamination in Out-of-Distribution (OOD) evaluation. Since many emulators are trained on emulated data from physics-based simulations, the models may inadvertently encounter future forcing patterns during pre-training. If an OOD test set contains scenarios that were part of the broader simulation pool used for model development, the evaluation is effectively in-distribution. In this work, we address this by strictly isolating a historical-only training regime to test true ``no-analog'' extrapolation.}

Standard ML techniques assume stationary data distributions, an assumption violated by climate trajectories characterized by nonlinear feedbacks and changing boundary conditions. Importantly, climate-change--driven shifts constitute not only differences in the marginal distribution $P(X)$, but also in the underlying physical relationships $P(Y|X)$. In such settings, models that have learned spurious correlations or historically bounded patterns {may experience degraded performance} when confronted with novel forcing combinations \citep{TS-OOG, CERA}. 

Despite rapid progress, current evaluation protocols provide limited insight into OOD robustness. Frameworks such as ClimateBench \citep{https://doi.org/10.1029/2021MS002954} typically evaluate models under a single emission scenario, leaving open the question of how these models behave when forced to extrapolate across divergent pathways. To address this gap, we introduce a comprehensive OOD evaluation framework tailored to the ``zero-shot'' extrapolation limits of ML climate emulators {trained exclusively on historical simulations (1850--2014) following the climate projection protocol established by Nguyen et al. \citep{nguyen2023climaxfoundationmodelweather}}. We benchmark three architectures (U-Net, ConvLSTM, ClimaX) across two primary axes of non-stationarity:
\begin{enumerate}
    \item \textbf{Temporal Extrapolation:} Testing models trained on historical forcing (1850--2014) against the recent, accelerated warming of 2015--2023.
    \item \textbf{Cross-Scenario Forcing Shifts:} Evaluating the capacity of historical-only models to predict divergent future pathways (SSP1--2.6 and SSP5--8.5).
\end{enumerate}

This work is guided by two core questions: (1) how do current ML climate emulators perform when restricted to historical training dynamics, and (2) what do these shifts reveal about the {performance-stability trade-off} between absolute predictive skill and {relative robustness under distribution change}? 

By systematically analyzing model robustness, we identify {an ``Accuracy vs. Stability'' pattern within this experimental configuration}: while foundation models achieve high absolute accuracy, they exhibit {larger relative performance changes} than simpler convolutional models when forced into novel climate regimes. Our results demonstrate that while temperature extrapolation is relatively stable, cross-scenario shifts in precipitation lead to substantial degradation {in the evaluated models}. 

This paper makes three main contributions:
\begin{itemize}
    \item We develop a unified OOD evaluation framework for ML-based climate emulation, integrating temporal extrapolation and cross-scenario forcing shifts to capture key forms of non-stationarity.
    \item We provide a systematic comparison of state-of-the-art architectures, {including a foundation model specifically evaluated under a historical-only training regime}, across these OOD dimensions.
    \item We demonstrate that cross-scenario forcing shifts induce substantial and model-dependent degradation, underscoring the need for domain generalization methods and {careful consideration of data contamination} in future emulator development.
\end{itemize}

\section{Methodology}

This section outlines the methodology for adapting Out-of-Distribution (OOD) evaluation techniques to address the challenges specific to ML climate modeling under non-stationary regimes. Our approach operationalizes the theoretical challenges of causal shifts and domain generalization by building on existing OOD frameworks in ML and tailoring them to climate-specific contexts, evaluating how ML models generalize across shifts in spatio-temporal climate data.

\subsection{Modeling OOD Generalization in Non-Stationary Climate Dynamics}

In traditional OOD settings, a model trained on source distribution $P_{\text{train}}$ struggles to generalize effectively to a target distribution $P_{\text{test}}$ where $P_{\text{train}} \neq P_{\text{test}}$. For high-dimensional, spatio-temporal data like climate variables, shifts are complex and can manifest across multiple domains.

Climate data is modeled as a time-series process, where observations $\mathbf{X}_t$ (e.g., climate forcers) lead to climate states $\mathbf{Y}_t$ (e.g., temperature and precipitation) at time $t$. Given a model $f$, the objective is to learn a mapping $f(\mathbf{X}_t) \approx \mathbf{Y}_{t+k}$ (for a $k$-step future state).

{In the context of ML climate emulation, a fundamental challenge arises from \textbf{data contamination}, where models are trained on simulated data from physics-based Earth System Models (ESMs) \citep{kendon2025potential}. Because ESM datasets often include a broad range of forcing scenarios, a model may inadvertently encounter physical patterns during pre-training that characterize future climate states. In such cases, what appears to be an OOD evaluation is actually in-distribution memorization of the ESM's internal physics. To address this, we strictly isolate the historical training window to test true ``no-analog'' extrapolation, forcing the model to encounter regime transitions and conditional shifts $P(\mathbf{Y}|\mathbf{X})$ that do not exist in its training history \citep{lutjens2025impact}.}

\subsubsection{Applying the Domain Generalization Framework}

We adapt the WOODS framework \citep{woods} for time-series evaluation, which systematically categorizes distribution shifts. We define a climate domain $d$ as a specific forcing scenario or climate regime, characterized by its joint distribution $P_d(\mathbf{X}_t, \mathbf{Y}_t)$.

The objective of Domain Generalization (DG) is to minimize the model's worst-case risk across all possible domains $E_{\text{all}}$, ensuring the learned predictor $f$ is robust to domain-specific shifts:
\[
\min_f \max_{d \in E_{\text{all}}} R_d(f)
\]
where the risk $R_d(f)$ for domain $d$ is the expected loss over the domain's distribution:
\[
R_d(f) = \mathbb{E}_{(\mathbf{X}_t, \mathbf{Y}_t)\sim P_d(\mathbf{X}, \mathbf{Y})} [\ell(f(\mathbf{X}_t), \mathbf{Y}_t)]
\]

WOODS \citep{woods} highlights two key scenarios for distribution shifts in time-series data:
\begin{enumerate}

    \item \textbf{Source-Domain Shifts:} Occur when training data comes from a domain that differs in underlying factors or conditions, such as different data sources.

    \item \textbf{Time-Domain Shifts:} Occur when the data distribution changes over time, which can be due to long-term trends, seasonal variations, or sudden, unexpected events.

\end{enumerate}

\subsubsection{Climate-Specific Distribution Shifts}

This distribution-shift taxonomy is highly relevant to ML-based climate modelling, where both source-domain and time-domain shifts arise naturally from the evolving Earth system. In climate applications, these shifts correspond to physically meaningful departures from the conditions represented in the training data:

\begin{itemize}
    \item \textbf{Source-Domain Shifts (Cross-Scenario Forcing Shifts):}  
    Source-domain shifts occur when models are trained under one set of external climate drivers but evaluated under another. In this work, these drivers correspond to the \emph{Shared Socioeconomic Pathways} (SSPs), which define standardized trajectories of greenhouse gas and aerosol emissions (e.g., $\mathrm{CO}_2$, $\mathrm{CH}_4$, $\mathrm{SO}_2$, black carbon), land-use patterns, and socioeconomic change. Different SSPs therefore encode distinct radiative forcing histories and futures, leading to divergent climate responses. As a result, cross-scenario evaluation probes not only shifts in the marginal distribution $P(X)$, but also differences in the underlying physical relationships $P(Y|X)$.  Testing emulators across SSP1--2.6 and SSP5--8.5 provides a direct measure of their robustness to fundamentally different causal forcing mechanisms {and reveals whether models have learned the underlying sensitivity or merely historically bounded correlations \citep{addison2024machine}}.

    \item \textbf{Time-Domain Shifts (Temporal Extrapolation):}  
    Temporal shifts arise when a model is applied to future periods that exhibit different dynamical behaviour than the historical data used for training. Long-term warming trends, changes in the frequency of extreme events, and the emergence of new climate regimes can all produce temporal drift between historical and future distributions. This evaluation assesses whether an emulator can \emph{extrapolate} into increasingly non-stationary climate states, even when future conditions fall outside the range of historical variability {\citep{rampal2024enhancing}}.
\end{itemize}

\subsection{OOD Evaluation Benchmarks}

We define three distinct and complementary OOD evaluation strategies leveraging the Latitude-Longitude Weighted Root Mean Squared Error (LL-RMSE) as the primary metric \citep{kaltenborn2023climatesetlargescaleclimatemodel}. 

\subsubsection{\textbf{Benchmark 1: Temporal Extrapolation (Time-Domain Shift)}}

This benchmark evaluates a model's ability to perform \emph{temporal extrapolation}. Models are trained exclusively on historical data (1850–2014) and evaluated on the subsequent recent period (2015–2023). This split is climate-relevant because the test window includes some of the warmest years on record and reflects a climate state increasingly influenced by anthropogenic forcing. As a result, the model must predict conditions that are systematically more extreme—and thus out-of-distribution—relative to the historical training regime. Temporal extrapolation therefore tests whether ML emulators can remain robust as the climate system evolves beyond the range of historical variability.

\subsubsection{\textbf{Benchmark 2: Forcing Pathways Robustness Analysis (Source-Domain Shift)}}

This benchmark assesses generalization across distinct \emph{forcing scenarios}. Using the SSPs, we systematically rotate the test set across two qualitatively different trajectories: SSP1–2.6 and SSP5–8.5. These scenarios were specifically selected to represent the most extreme plausible futures: SSP1–2.6 acts as the optimistic "sustainability" pathway with aggressive mitigation, while SSP5–8.5 represents the high-emissions "fossil-fueled development" extreme. Each SSP encodes a different radiative forcing pathway and socio-economic development trajectory, resulting in different climate responses. Cross-scenario evaluation therefore probes model robustness under shifts in the causal mechanisms driving climate evolution. Unlike existing benchmarks that evaluate models under a single scenario, this method provides a more rigorous assessment of how well models generalize across diverse future forcing pathways.

\subsubsection{\textbf{Other Potential Future OOD Benchmarks}}

A comprehensive assessment of OOD generalisation in climate models can be approached through several complementary dimensions of distribution shift. Beyond temporal shifts and cross-scenario forcing changes, OOD robustness may be evaluated through spatial-domain shifts—testing models trained in one climate regime (e.g., midlatitudes) on another (e.g., tropics)—and resolution shifts, where models must generalise from coarse-grid training data to finer spatial structures. Furthermore, a systematic approach could involve the automated detection of extreme or out-of-distribution events within existing datasets; by isolating these high-anomaly samples, researchers could partition data into a refined "in-distribution" training set and a specifically targeted, challenging OOD test set. Structural-domain shifts, in which models trained on one family of GCMs are tested on structurally distinct models, probe sensitivity to climate-model formulation and parameterisation. Additional axes include internal-variability shifts driven by regime changes such as ENSO phases, counterfactual or intervention-style perturbations (e.g., volcanic aerosol pulses) designed to test extrapolation beyond historically observed forcings, and extreme-event tail evaluations that quantify robustness under rare but high-impact anomalies. Finally, physics-consistency checks and extended forecast-horizon tests can reveal unphysical behaviour or cumulative drift when models are used beyond their training regimes. Together, these evaluation modes could provide a richer and more realistic characterisation of how ML climate emulators may fail—or remain reliable—under the deeply non-stationary dynamics of a changing climate.

\subsection{Dataset, Models, and Evaluation Metrics} 

\begin{itemize}
    \item \textbf{Dataset:} We utilize the ClimateSet dataset \citep{kaltenborn2023climatesetlargescaleclimatemodel}, comprising monthly forcing agents ($CO_2, CH_4, BC, SO_2$) and outputs ($tas, pr$) from CMIP6.
    \item \textbf{Models:} We evaluate three architectures representing diverse spatio-temporal inductive biases: \textbf{U-Net} (convolutional encoder-decoder), \textbf{ConvLSTM} (recurrent-convolutional), and \textbf{ClimaX} (transformer-based foundation model).
    \item \textbf{{ClimaX Training Protocol:}} {We follow the climate projection task protocol of the original ClimaX paper \citep{nguyen2023climaxfoundationmodelweather}. ClimaX is initialized with weights pre-trained on CMIP6 historical simulations. For our experiments, the model is fine-tuned to map the four input forcing agents to the atmospheric response variables. }
    \item \textbf{Evaluation Metrics:} Performance is quantified using LL-RMSE to account for grid-cell area variation at high latitudes.
\end{itemize}

\subsection{Experimental Setup for Distribution Shifts} 

Our setup introduces controlled modifications to the training and testing domains to create ``zero-shot'' extrapolation challenges. To ensure a ``like-for-like'' comparison, we define our In-Distribution (ID) baseline as the performance on a held-out historical test set. This ensures the baseline represents the model's peak performance under the exact distribution it was trained on.

\subsubsection{Data Partitioning and Splits}
{To ensure a rigorous OOD test and avoid the contamination issues discussed before \citep{kendon2025potential}, we implement a temporal split. The training and ID validation sets are restricted to 1850--2014. No future scenario data or post-2014 observations are included in any part of the training or hyperparameter selection process. This setup ensures that the benchmarks in Table \ref{tab:data_splits} represent true ``zero-shot'' extrapolation challenges.}

\begin{table}[ht]
\centering
\caption{Data Partitioning for ML Training and OOD Benchmarking}
\label{tab:data_splits}
\begin{tabular}{@{}llll@{}}
\toprule
\textbf{Dataset Role} & \textbf{Scenario(s)} & \textbf{Year Range} & \textbf{Split Strategy} \\ \midrule
Training (Train) & Historical & 1850--2014 & 90\% random samples \\
In-Distribution (ID Test) & Historical & 1850--2014 & 10\% held-out random samples \\
Benchmark 1 (OOD Test) & Historical & 2015--2023 & Time-Domain Shift \\
Benchmark 2 (OOD Test) & SSP1-2.6, SSP5-8.5 & 2015--2100 & Source-Domain Shift \\ \bottomrule
\end{tabular}
\end{table}

\subsubsection{Consistent Initialization and Benchmarking}
We adopted the standardized training parameters from ClimateSet \citep{kaltenborn2023climatesetlargescaleclimatemodel}, using 50 epochs with learning rates of $2 \times 10^{-4}$ for U-Net/ConvLSTM and $5 \times 10^{-4}$ for ClimaX. 

\subsubsection{Quantifying Model Sensitivity}
We quantify shift impact using \textbf{Percent Change} relative to the ID baseline:
\begin{equation}
\text{Percent Change} = \frac{ \text{LL-RMSE}_{\text{shift}} - \text{LL-RMSE}_{\text{baseline}} }{ \text{LL-RMSE}_{\text{baseline}} } \times 100
\end{equation}
{This metric isolates the model's sensitivity to deviations from historical dynamics, providing a directional indicator of robustness that is independent of absolute error magnitudes.}
{While absolute LL-RMSE varies across scenarios and variables, this relative measure provides a transparent diagnostic framework for assessing directional sensitivity as models encounter novel climate states.}

\subsubsection{{Methodological Challenges: Data Contamination and Domain Overlap}}

{A significant challenge in evaluating the OOD robustness of climate foundation models is the potential for data contamination. Because ClimaX was pre-trained on the broad CMIP6 historical archive \citep{nguyen2023climaxfoundationmodelweather}, the model has been previously exposed to the underlying physical parameterizations of the Earth System Models used in our test sets. This creates a "blurred boundary" between in-distribution and OOD data, as the model may rely on memorized physical patterns rather than true zero-shot generalization.}

{Our decision to enforce a historical-only training regime (1850--2014) for the task-specific prediction head is a deliberate methodological choice to mitigate this overlap. By isolating the model from any future scenario data (SSPs) during the fine-tuning stage, we force the architecture to extrapolate to no-analog conditions. Consequently, the performance metrics reported in this work should be interpreted as indicators of the model's directional sensitivity to non-stationary shifts, providing a more conservative and realistic assessment of how such systems might behave when confronted with truly unprecedented climate states.}

\section[Results and Analysis]{Results and Analysis}

The performance of the evaluated ML models varies across our two OOD benchmarks, but a consistent trend of high absolute error is visible across all tests, including the in-distribution (ID) baseline. Before analyzing the relative shifts in performance, we must acknowledge the absolute error levels as a primary limitation of this study.

The absolute LL-RMSE values achieved (ranging from 0.8 to 1.1) represent significantly lower accuracy than state-of-the-art emulators, which typically achieve errors in the 0.2--0.3 range. Notably, this performance floor persists even on the in-distribution test sets. We identify three primary factors contributing to this underfitted state:

\begin{itemize}
\item \textbf{Standardized Training vs. Peak Optimization:} To ensure a consistent and comparable evaluation across all architectures, we adopted the standardized training parameters established in the ClimateSet benchmark \citep{kaltenborn2023climatesetlargescaleclimatemodel}. {Specifically, our ClimaX implementation follows the climate projection task protocol described in Section 4.3 of the original ClimaX paper \citep{nguyen2023climaxfoundationmodelweather}, where the model is fine-tuned to map forcing agents to atmospheric responses using a task-specific prediction head.} While this configuration-control approach prevents per-scenario hyperparameter tuning from masking intrinsic architectural fragility, it necessarily results in sub-optimal absolute performance compared to extensively tuned emulators.

\item \textbf{Historical Data Scarcity and Anchoring:} The historical period (1850--2014) is characterized by lower variance in climate forcing agents compared to future scenarios. Consequently, the models were ``anchored'' to a relatively stable climate regime, making it difficult for them to learn the high-sensitivity dynamics required to capture the more extreme climate states encountered in the 21st-century test sets. {This anchoring effect is a consequence of our decision to enforce a strict historical isolation protocol, which aims to test true extrapolation without the benefit of seeing future forcing trends during the fine-tuning stage.}

\item \textbf{Architectural Sensitivity:} The high error in the ID test set suggests that without extensive, task-specific tuning, current architectures struggle to capture the complex, multi-scale spatial dependencies of variables like precipitation from historical records alone.
\end{itemize}

Despite these high absolute errors, the study’s objective remains to quantify relative sensitivity. By maintaining a strictly consistent training regime, we isolate the impact of the domain shift and reveal how different inductive biases respond when moving from ID data to novel OOD scenarios.

\subsection*{Architectural Sensitivity and the ClimaX Paradox}

Within this baseline of high absolute error, distinct architectural behaviors emerge. While ClimaX consistently achieved the lowest absolute LL-RMSE scores across all experiments---maintaining its position as the most accurate architecture in our study---it also exhibited the highest sensitivity to OOD shifts. Its error metrics fluctuated significantly more than the CNN-based baselines, U-Net and ConvLSTM, when moving from in-distribution to OOD data.

We hypothesize that ClimaX’s performance in this context may represent an optimistic upper bound. {A key methodological consideration is the potential for data contamination and domain overlap; because ClimaX was pretrained on a vast collection of CMIP6 historical data \citep{nguyen2023climaxfoundationmodelweather}, the backbone possessed prior exposure to the fundamental physical parameterizations of the Earth System Models used in our test sets.} {This creates a "blurred boundary" between ID and OOD evaluation, as the model may rely on memorized representations from pre-training rather than learned physical generalization from the fine-tuning data.} The fact that ClimaX still exhibited the highest relative performance degradation despite this {prior exposure} suggests that the Transformer's global attention mechanism may be {more sensitive to shifts in forcing magnitudes} than the local spatial priors of CNNs when operating in {non-stationary} or ``no-analog'' environments.

\subsection*{Benchmark 1 Results - Time-Domain Shift}

The historical training period (1850–2014) and the testing period (2015–2023) were separated to evaluate the models’ ability to generalize across time. {Following the protocol in Section 4.3 of Nguyen et al. (2023), ClimaX was fine-tuned specifically on the 1850--2014 window. Since its foundational pre-training also utilized data up to 2015 \citep{nguyen2023climaxfoundationmodelweather}, this test period reflects a boundary case where the model encounters a regime shift that was minimally present in its pre-training but entirely absent from its fine-tuning data.}

\begin{figure}[!htb]
\centering
\includegraphics[width=\textwidth]{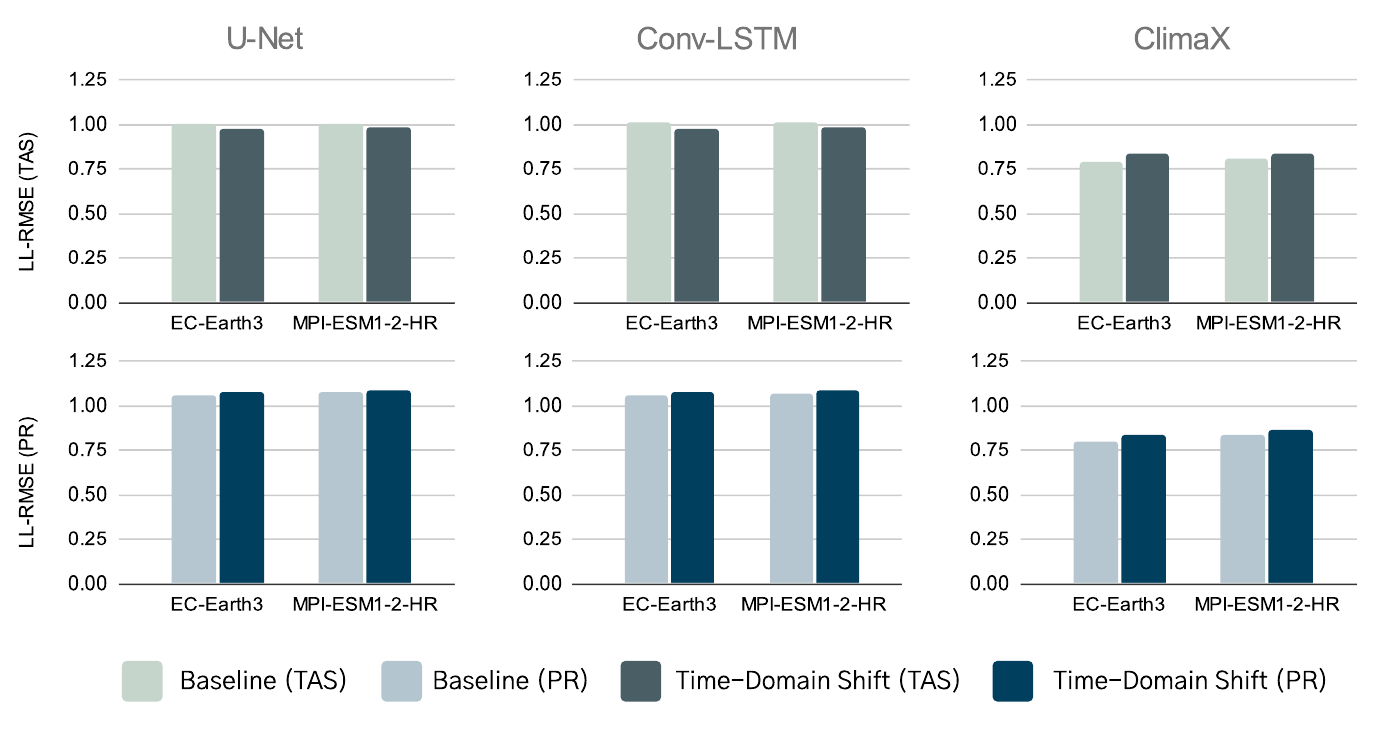}
\caption{Comparison of surface air temperature (TAS) and precipitation (PR) test LL-RMSE for ML models across Baseline and temporal shift test sets.}
\label{fig:method1_tas_pr}
\end{figure}

The results, shown in Table \ref{tab:percent_change_temporal} and Figure \ref{fig:method1_tas_pr}, reveal a split in performance between the two variables. For Temperature (TAS), UNet and Conv-LSTM achieved moderate negative percent changes in LL-RMSE (ranging from -1.2\% to -3.5\%). {This result—where models show a lower relative error on the OOD window than on the historical baseline—suggests that the 2015–2023 period may represent a "simpler" emulation regime for temperature.} This is likely due to the strong, monotonic warming trend driven by anthropogenic forcing in the last decade, which is easier for ML models to emulate than the higher inter-decadal variability present in the earlier historical record.

In contrast, Precipitation (PR) saw a consistent performance degradation across all architectures, with LL-RMSE increasing by up to 5.18\%. Unlike the thermodynamic certainty of temperature, precipitation is governed by complex, threshold-based physics and local moisture convergence. Even a short 9-year temporal shift introduces enough atmospheric {variability} to degrade the performance of models that lack explicit hydrological constraints.

Notably, while ClimaX achieved the lowest absolute LL-RMSE scores, it was the most sensitive to the temporal shift. Its temperature error increased by 4.8\% and 2.6\% across the two GCMs, while other models {showed relative improvements}. {This suggests a measurable tension: while foundation models like ClimaX offer superior baseline accuracy, their global attention mechanism may be more susceptible to performance "drift" as the climate state evolves, particularly when the task-specific prediction head is restricted to historical training.}

\begin{table}[ht!]\label{Table1}
\tabcolsep=0pt%
\caption{Percent change (\%) for ML Models emulating surface air temperature (TAS) and precipitation (PR) under temporal shift benchmark.\label{tab:percent_change_temporal}}
\begin{tabular*}{\textwidth}{@{\extracolsep{\fill}}lcccccc@{}}\toprule%
 & \multicolumn{2}{@{}c@{}}{{U-Net}} 
 & \multicolumn{2}{@{}c@{}}{{Conv-LSTM}} 
 & \multicolumn{2}{@{}c@{}}{{ClimaX}} \\
\cmidrule{2-3} \cmidrule{4-5} \cmidrule{6-7}%
{Model} 
& {TAS} & {PR} 
& {TAS} & {PR} 
& {TAS} & {PR} \\\midrule
EC-Earth3 
& -2.90 & 2.31 
& -3.55 & 1.92 
& 4.82 & 5.18 \\
MPI-ESM1-2-HR 
& -1.28 & 1.59 
& -1.97 & 0.43 
& 2.61 & 2.83 \\
\end{tabular*}%
\end{table}

\subsection*{Benchmark 2 Results - Source-Domain Shift}

\begin{figure}[!htb]
\centering
\includegraphics[width=\textwidth]{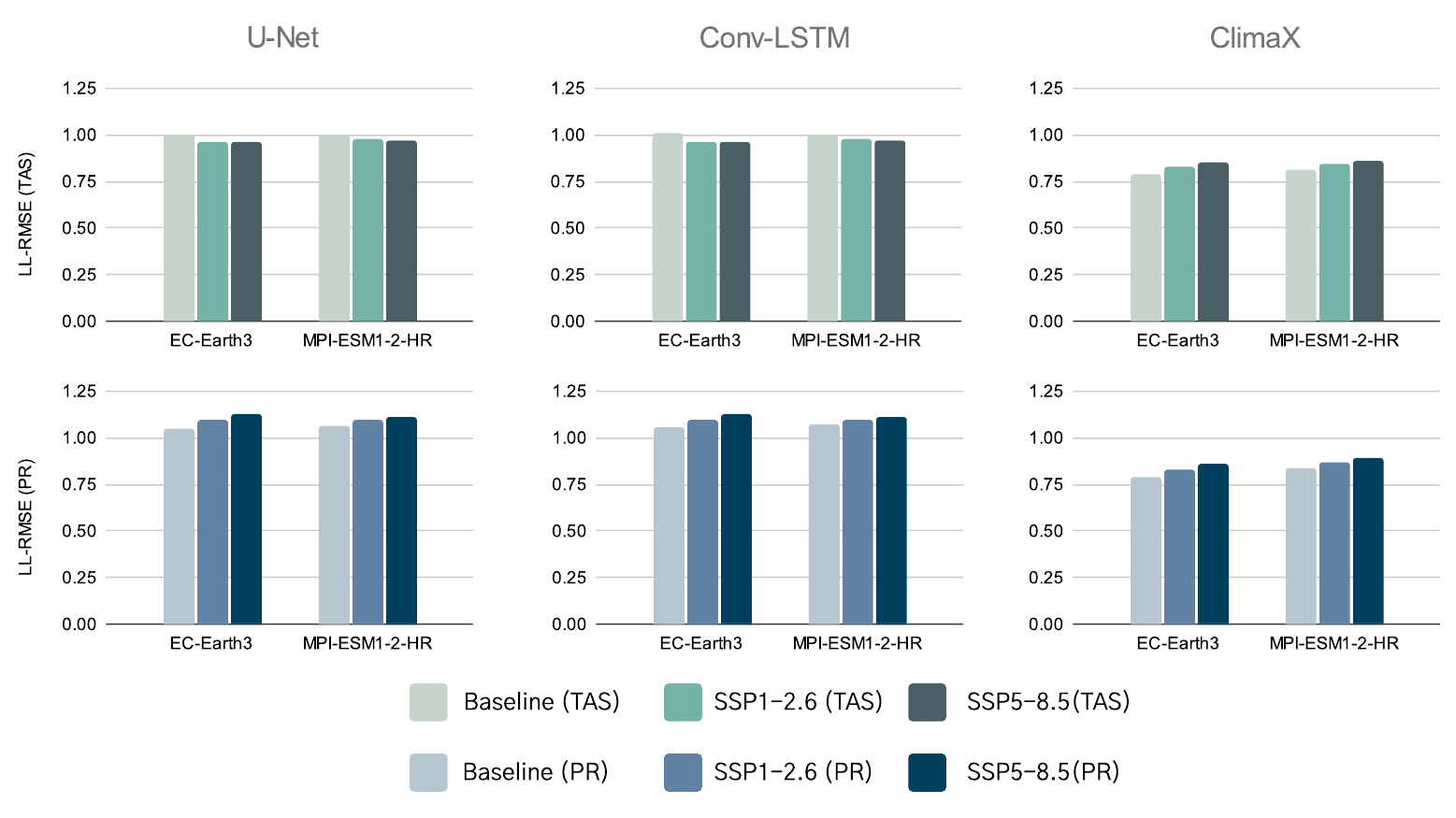}
\caption{Comparison of surface air temperature (TAS) and precipitation (PR) test LL-RMSE for ML models across Baseline, SSP1-2.6 and SSP5-8.5 test sets.}
\label{fig:method2_tas_pr}
\end{figure}

This experiment tested the capacity of models trained exclusively on historical data to predict unseen future forcing pathways: SSP1-2.6 and SSP5-8.5. {The results, detailed in Table \ref{tab:percent_change} and Figure \ref{fig:method2_tas_pr}, reveal a measurable shift in model reliability across variables under divergent forcing trajectories.}

For surface air temperature (TAS), models generally maintained or even improved their LL-RMSE relative to the historical baseline. This reflects the strong, linear thermodynamic coupling between $CO_2$ concentration and global warming. However, ClimaX proved to be an outlier; despite its high baseline accuracy, its error increased by up to 8.18\% in SSP5-8.5. {This heightened sensitivity suggests that ClimaX may be more "anchored" to the historical climate state seen during pre-training, leading to a higher relative increase in error when faced with the intensified thermal gradients of a high-emissions future.}

The results for precipitation (PR) signal a {significant challenge for robustness}. Across all architectures and GCMs, performance degraded significantly, with error increases scaling alongside the severity of the forcing. In the extreme SSP5-8.5 scenario, precipitation errors rose by as much as 8.44\%. {This consistent pattern of degradation indicates that the evaluated architectures struggle to capture the non-linear, stochastic responses of the hydrological cycle when pushed into "no-analog" climate states.}

\begin{table}[ht!]\label{Table2}
\tabcolsep=0pt%
\caption{Percent change (\%) for ML Models emulating surface air temperature (TAS) and precipitation (PR) under SSP1-2.6 scenario test set\label{tab:percent_change}}
\begin{tabular*}{\textwidth}{@{\extracolsep{\fill}}lcccccc@{}}\toprule%
 \textbf{SSP1-2.6} & \multicolumn{2}{@{}c@{}}{{U-Net}} 
 & \multicolumn{2}{@{}c@{}}{{Conv-LSTM}} 
 & \multicolumn{2}{@{}c@{}}{{ClimaX}} \\
\cmidrule{2-3} \cmidrule{4-5} \cmidrule{6-7}%
{Model} 
& {TAS} & {PR} 
& {TAS} & {PR} 
& {TAS} & {PR} \\\midrule
EC-Earth3 
& -3.58 & 4.18 
& -4.23 & 3.77 
& 4.61 & 4.76 \\
MPI-ESM1-2-HR 
& -1.92 & 3.05 
& -2.61 & 1.88 
& 4.18 & 3.99 \\
\bottomrule
 \textbf{SSP5-8.5} & \multicolumn{2}{@{}c@{}}{{U-Net}} 
 & \multicolumn{2}{@{}c@{}}{{Conv-LSTM}} 
 & \multicolumn{2}{@{}c@{}}{{ClimaX}} \\
\cmidrule{2-3} \cmidrule{4-5} \cmidrule{6-7}%
{Model} 
& {TAS} & {PR} 
& {TAS} & {PR} 
& {TAS} & {PR} \\\midrule
EC-Earth3 
& -3.92 & 7.15 
& -4.57 & 6.74 
& 8.18 & 8.44 \\
MPI-ESM1-2-HR 
& -2.69 & 4.72 
& -3.37 & 3.53 
& 6.27 & 6.85 \\
\end{tabular*}%
\end{table}

\subsection*{Results Summary}

Our benchmarks reveal that the efficacy of machine learning emulators is highly sensitive to the nature of the target variable and the severity of the distribution shift. {These findings suggest that increasing architectural complexity—specifically moving from convolutional structures to global Transformer-based models—does not necessarily improve relative robustness in non-stationary environments.}

\begin{itemize}
\item \textbf{Thermodynamic Stability vs. Hydrological Stochasticity:} Models successfully extrapolate temperature (TAS) trends, {benefiting from the high signal-to-noise ratio in anthropogenically forced warming}. However, they struggle significantly with precipitation (PR), where the stochastic nature of the hydrological cycle leads to {measurable performance degradation}. {This suggests that judged solely by temperature performance, emulators may appear more robust than they are when applied to variables critical for local climate impacts.}

\item \textbf{The Accuracy-Stability Tension:} {We observe an identifiable pattern of performance variation:}
\begin{itemize}
\item CNN-based models (U-Net, Conv-LSTM) utilize local spatial priors that {exhibited lower relative volatility} when forced to generalize into "no-analog" high-emission futures (SSP5-8.5).
\item ClimaX, while achieving the highest absolute accuracy in-distribution, exhibits the greatest {relative sensitivity} to domain shifts. {As noted in Section 3, this may be exacerbated by the "data contamination" of its foundational pre-training, which creates a blurred boundary between interpolation and true OOD extrapolation.}
\end{itemize}
\end{itemize}

{These insights underscore that model performance in climate emulation may be task-dependent: while complexity provides better fits for historical interpolation, the simpler architectures with local spatial priors evaluated here offered more stable relative performance for long-term climate projections.}

\section{Conclusion}

This study assessed the robustness of three state-of-the-art ML-based global climate emulators---U-Net, ConvLSTM, and ClimaX---under a comprehensive suite of out-of-distribution (OOD) conditions. By introducing an evaluation framework that operationalizes temporal shifts and cross-scenario forcing shifts, we demonstrated that while these architectures are increasingly viable for near-term climate tasks, they remain sensitive to the physical distributions of their training data. This sensitivity is particularly pronounced in precipitation emulation, which shows substantial degradation under forcing-pathway changes, suggesting that current models are limited in their ability to extrapolate to ``no-analog'' climate regimes.

{The performance gaps identified here suggest that while ML emulators are powerful tools for accelerating climate projections and analyzing potential extreme events, their deployment carries inherent risks if their OOD robustness is not fully understood. In many high-stakes applications—such as disaster preparedness or long-term infrastructure planning—an emulator’s high absolute accuracy on historical data may provide a false sense of security. If a model fails to generalize under no-analog conditions, it may produce misleading projections of extreme events precisely when accurate information is most critical. This underscores the need for more rigorous approaches to OOD testing that move beyond standard validation metrics and focus on characterizing the limits of physical extrapolation.}

{A fundamental hurdle in this endeavor is the problem of data contamination. Because many climate foundation models are pre-trained on emulated data from physics-based simulations that already encompass future scenarios, true OOD evaluation is often masked by the model’s prior exposure to the test distribution \citep{kendon2025potential}. By enforcing a strict historical-isolation protocol, this work reveals that when restricted to historical dynamics, models may struggle to capture the non-linear physics of a rapidly warming world. To mitigate these risks, future research must prioritize ``scenario-aware'' training—incorporating diverse and divergent future forcing pathways to break the historical anchor and improve extrapolation stability \citep{lutjens2025impact, rampal2024enhancing}. Only by disentangling external forcing shifts from internal variability and investigating causal representation techniques can we develop emulators that are not only computationally efficient but also physically robust under the non-stationary conditions of the 21st century.}

\bibliographystyle{apalike}
\bibliography{bibliography}

\begin{appendix}
\section{Appendix }\label{appendixA}

\begin{table}[hbt!]
\caption{Baseline and Temporal Shift LL-RMSE values for each ML model: surface air temperature (TAS) and precipitation (PR) predictions\label{tab:appendix_loss_comparison_6}}
\centering
\begin{tabular}{l|c|cc|cc|cc}
\toprule
Model & Scenario 
& \multicolumn{2}{c}{U-Net} 
& \multicolumn{2}{c}{Conv-LSTM} 
& \multicolumn{2}{c}{ClimaX} \\
\cmidrule{3-4} \cmidrule{5-6} \cmidrule{7-8}
      &          
& TAS & PR 
& TAS & PR 
& TAS & PR \\
\midrule
EC-Earth3 
& Time Shift 
& 0.971 & 1.075 
& 0.971 & 1.075 
& 0.829 & 0.834 \\
& Baseline 
& 1.000 & 1.051 
& 1.007 & 1.055 
& 0.791 & 0.793 \\
\midrule
MPI-ESM1-2-HR 
& Time Shift 
& 0.985 & 1.080 
& 0.985 & 1.080 
& 0.831 & 0.858 \\
& Baseline 
& 0.998 & 1.063 
& 1.005 & 1.075 
& 0.810 & 0.834 \\
\end{tabular}
\end{table}

\begin{table}[hbt!]
\caption{Baseline and SSP1-2.6 LL-RMSE values for each ML model, surface air temperature (TAS) and precipitation (PR) predictions\label{tab:appendix_loss_comparison}}
\centering
\begin{tabular}{l|c|cc|cc|cc}
\toprule
Model & Scenario
& \multicolumn{2}{c}{U-Net}
& \multicolumn{2}{c}{Conv-LSTM}
& \multicolumn{2}{c}{ClimaX} \\
\cmidrule{3-4} \cmidrule{5-6} \cmidrule{7-8}
      &
& TAS & PR
& TAS & PR
& TAS & PR \\
\midrule
EC-Earth3
& SSP1-2.6
& 0.965 & 1.095
& 0.965 & 1.095
& 0.828 & 0.831 \\
& Baseline
& 1.000 & 1.051
& 1.007 & 1.055
& 0.791 & 0.793 \\
\midrule
MPI-ESM1-2-HR
& SSP1-2.6
& 0.979 & 1.095
& 0.979 & 1.095
& 0.844 & 0.868 \\
& Baseline
& 0.998 & 1.063
& 1.005 & 1.075
& 0.810 & 0.834 \\
\end{tabular}
\end{table}

\begin{table}[hbt!]
\caption{Baseline and SSP5-8.5 LL-RMSE values for each ML model: surface air temperature (TAS) and precipitation (PR) predictions\label{tab:appendix_loss_comparison_2}}
\centering
\begin{tabular}{l|c|cc|cc|cc}
\toprule
Model & Scenario
& \multicolumn{2}{c}{U-Net}
& \multicolumn{2}{c}{Conv-LSTM}
& \multicolumn{2}{c}{ClimaX} \\
\cmidrule{3-4} \cmidrule{5-6} \cmidrule{7-8}
      &
& TAS & PR
& TAS & PR
& TAS & PR \\
\midrule
EC-Earth3
& SSP5-8.5
& 0.961 & 1.126
& 0.961 & 1.126
& 0.856 & 0.860 \\
& Baseline
& 1.000 & 1.051
& 1.007 & 1.055
& 0.791 & 0.793 \\
\midrule
MPI-ESM1-2-HR
& SSP5-8.5
& 0.971 & 1.113
& 0.971 & 1.113
& 0.861 & 0.892 \\
& Baseline
& 0.998 & 1.063
& 1.005 & 1.075
& 0.810 & 0.834 \\
\end{tabular}
\end{table}

\end{appendix}

\end{document}